\documentclass{article}
\usepackage{spconf,amsmath,graphicx, enumitem, verbatim, threeparttable}
\usepackage{multirow}
\usepackage{multicol}
\usepackage{makecell}
\usepackage[bottom]{footmisc}

\title{Visualizing and alleviating the effect of radial distortion on camera calibration using principal lines
}
\name{Jen-Hui Chuang and Hsin-Yi Chen}
\address{Department of Computer Science, National Yang Ming Chiao Tung University, Taiwan, ROC}
\begin{document}
%
\maketitle
\begin{abstract}
Preparing appropriate images for camera calibration is crucial to obtain accurate results. In this paper, new suggestions for preparing such data to alleviate the adverse effect of radial distortion for a calibration procedure using principal lines are developed through the investigations of: (i) identifying directions of checkerboard movements in
an image which will result in maximum (and minimum) influence on the calibration results, and (ii) inspecting symmetry and monotonicity of such effect in (i) using the above principal lines. Accordingly, it is suggested that the estimation of principal point should based on linearly independent pairs of nearly parallel principal lines, with a member in each pair corresponds to a near 180° rotation (in the image plane) of the other. Experimental results show that more robust and consistent calibration results for the foregoing estimation can actually be obtained, compared with the renowned algebraic methods which estimate distortion parameters explicitly.
\end{abstract}
\section{Introduction}
\label{sec:intro}

Camera calibration estimates intrinsic and extrinsic camera parameters to associate an image pixel to the corresponding 3D point in the real world. In general, accurate calibration is a necessity for accurate representation of the real world using information extracted from the captured images and thus play a crucial role in computer vision applications such as intelligent video surveillance \cite{Appli_video_surveillance}, autonomous driving systems \cite{Appli_auto_driving_01, Appli_auto_driving_02} and accurate measurements \cite{Appli_accurate_measurement}.\par

For more than two decades, Zhang’s method \cite{zhang2000,zhang1998} has been the most popular solution to camera calibration due to its simplicity of only using images of a checkerboard (CB), in various poses, to estimate all camera parameters. The method is based on an algebraic closed-form solution in getting an initial estimation, before a nonlinear optimization, which minimizes the re-projection error of corner points of the CB on the images, is performed to finalize the estimation of camera parameters including distortion coefficients. Incidentally, it is suggested in \cite{zhang2000} that the CB patterns should cover whole image, rather than just occupying a small part of the image, for more accurate estimate of distortion coefficients.\par

Under the foregoing condition of whole image coverage, corners of the CB pattern may still concentrate on the less distorted center part of the image plane. Accordingly, the idea is extended in \cite{distribute02} to suggest that each image should be completely occupied by (part of) the pattern, with essentially no background area left, for better estimation of camera distortion. As the CB is partially visible, special patterns are included in the new CB pattern to simplify the identification of its visible corner points. In addition, somehow similar suggestion is proposed in \cite{distribute01}, i.e., corner points should be distributed uniformly, on the average, for a set of CB images without considering the distortion explicitly.\par

Given a set of CB patterns, there are calibration methods which suggest additional CB poses for better calibration results.  Assuming Gaussian noises, the next best pose is directly estimated in \cite{estimate_next_01} by minimizing both the calibration and corner uncertainty. However, system errors from the distortion may cause the method to fail. Such errors are further decreased in \cite{estimate_next_02} by increasing parameters of the pinhole camera model but without guaranteeing low system errors. In \cite{estimate_next_03}, a score function is proposed to estimate an optimal pose set in offline mode. However, such scheme is unrealistic as some camera parameters, which are yet to be determined, are used in the score function. Besides, the method does not consider distortion explicitly either.\par
  
Unlike the above methods which are based on Zhang's algebraic formulation to estimate all camera parameters simultaneously. A novel geometry-based camera calibration is proposed in \cite{chuang2021} which estimated camera geometry sequentially by first finding principal lines (PLs). A PL of a CB, e.g., the $y$-axis shown in Fig.  \ref{fig:relationship_of_pimg_and_pchkb} (a), corresponds to the intersection of image plane and the plane containing camera optical axis and perpendicular to both the CB and the image. As a PL thus defined should pass through the principal point (PP), a set of linearly independent PLs are used in \cite{chuang2021} to find the PP as their intersection. Thus, PLs with problematic intersections can simply be identified as outliers without resorting to probabilistic RANSAC \cite{RANSAC_01, RANSAC_02}.\par

In this paper, variation of PP location resulted from different sets of PLs, each having the above property of symmetry, are used to visualize of the adverse
effect of radial distortion on camera calibration, while novel and systematic arrangements of CB poses are suggested accordingly for alleviating such effect. The main contributions of the paper include:
\begin{enumerate}[label=(\roman*)]
    \itemsep=-4pt
    \item  (Directionality analysis) Identifying directions of CB shift in the image which will result in maximum deflection of the PL due to radial distortion. \label{item:first}
    \item (Monotonicity analysis) Demonstrating that the deflection in \ref{item:first}, which can affect PP estimate, will increase monotonically with the amount of CB shift. \label{item:second}
    \item Providing suggestions of CB poses, which are easy to set up, for minimizing the effect of distortion. \label{item:third}
\end{enumerate}

In Sec. 2, the setup of a reference set of CB poses used in \cite{chuang2021} are reviewed, wherein near perfect PLs (and the PP) can be obtained without considering image distortion. In Sec. 3, variation of this reference is employed to visualize effect of the distortion, before \ref{item:third} is accomplished. Finally, some experimental results and concluding remarks are provided.

\section{Geometry of a reference set of CB poses used in [12]}
\label{sec:review}

\begin{figure}[htb]

\begin{minipage}[b]{1.0\linewidth}
  \centering
  \centerline{\includegraphics[width=5.5cm]{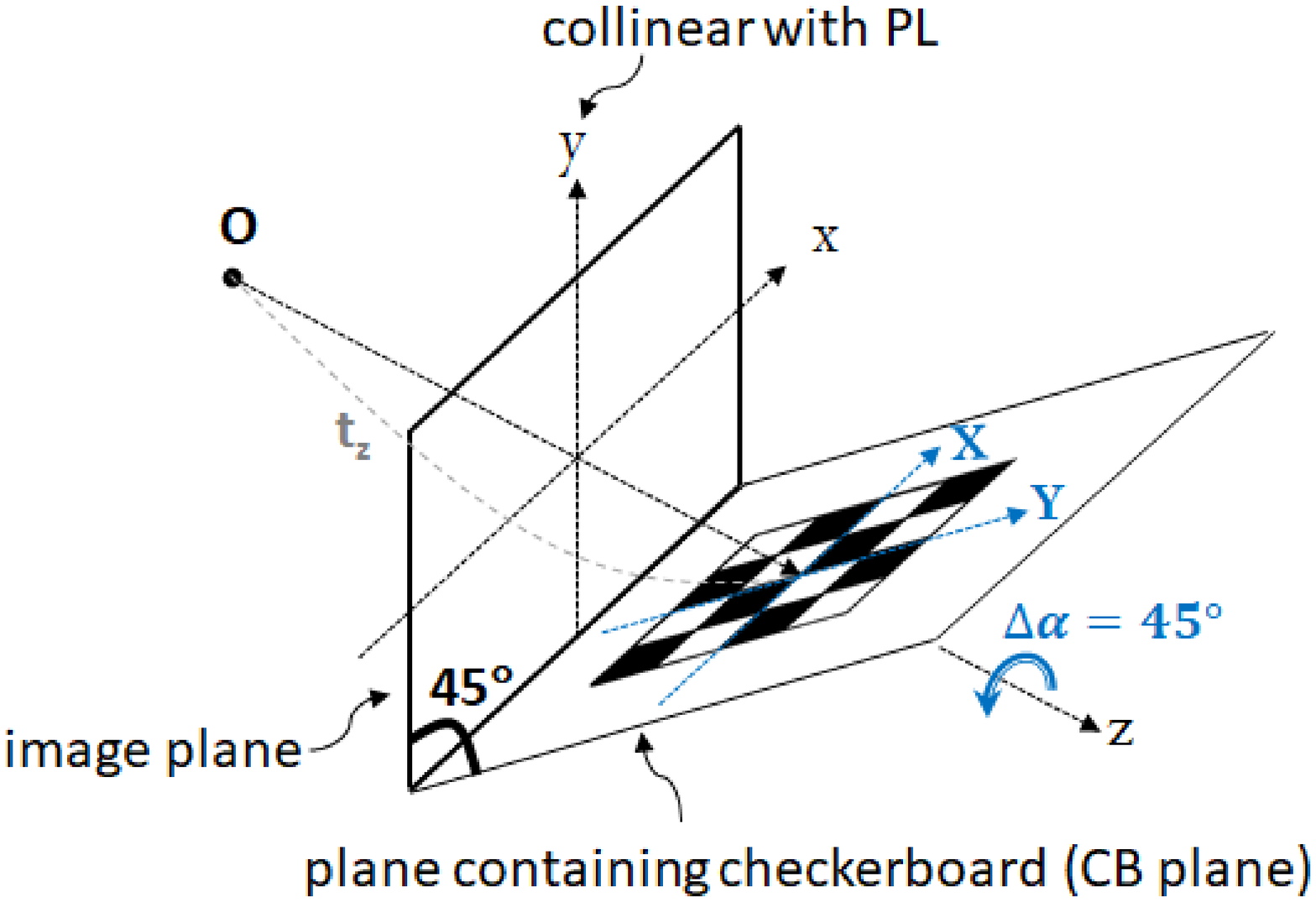}}
  \centerline{(a)}\medskip
\end{minipage}

\begin{minipage}[b]{1.0\linewidth}
  \centering
  \centerline{\includegraphics[width=7.0cm]{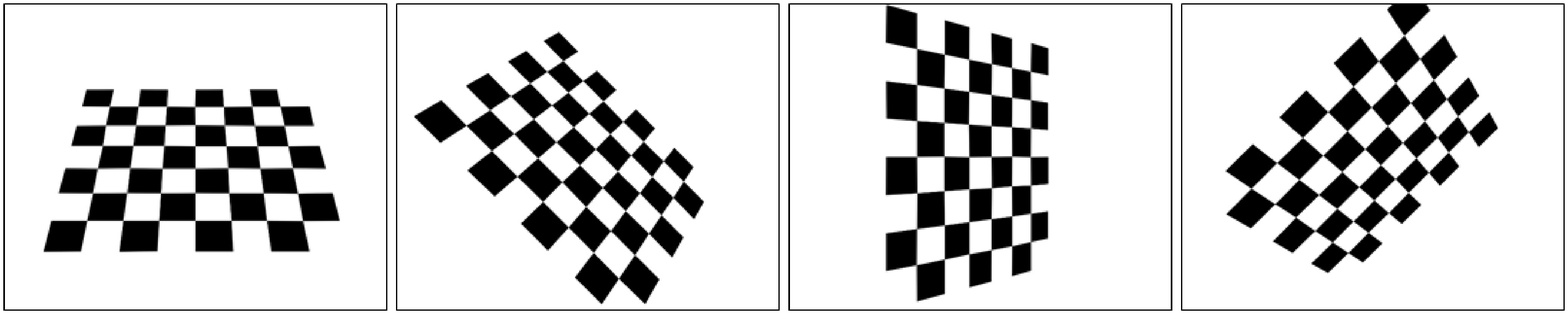}}
  \centerline{(b)}\medskip
\end{minipage}

\caption{(a) The dihedral angle between image plane and CB plane suggested in \cite{zhang1998, zhang2000, chuang2021}. (b) Four (of eight) CB images obtained from (a) according to \cite{chuang2021}.}

\label{fig:relationship_of_pimg_and_pchkb}

\end{figure}

In this section, a simple setup of CB poses employed in \cite{chuang2021} is reviewed, which is good for excellent estimate of PP, with all PLs having a near perfect intersection. Fig. \ref{fig:relationship_of_pimg_and_pchkb} (a) illustrates the geometry of such setup, with a dihedral angle of about 45° between image plane and plane containing CB (the CB plane) suggested in \cite{zhang1998, zhang2000, chuang2021} for better performance in camera calibration. In addition, it also is suggested in \cite{chuang2021} that eight linearly independent PLs can be used in the calibration by rotating the CB plane in Fig. \ref{fig:relationship_of_pimg_and_pchkb} (a) w.r.t. the z-axis (the optical axis), each time with $\Delta\alpha = 45^{\circ}$, as shown in Fig. \ref{fig:relationship_of_pimg_and_pchkb} (b) for four of them.\footnote{Lacking the geometry of PL, similar positions and orientations of CB are never suggested previously.} Note that the geometry in Fig. \ref{fig:relationship_of_pimg_and_pchkb} needs not be exact as the precise image coordinate system is unavailable during the calibration.

While better calibration results are obtained in \cite{chuang2021} compared with Zhang's method, the effect of distortion is actually considered in the latter but not in the former, which may due to the fact that the CB patterns are roughly located at the central region of the camera view where the distortion is minimal. In this paper, the effect of radial distortion will be examined closely by investigating such effect on the deflection of PL due to the outward translation of the CB pattern.

\section{Visualization of radial distortion}
\label{sec:visualization}

In this section, variation of the above reference poses will be employed to visualize the effect of camera distortion. Specifically, deflection of the PL due to different direction/amount of translation of the CB pattern shown in Fig. \ref{fig:relationship_of_pimg_and_pchkb} (a) is used for the visualization before appropriate placements of the pattern are suggested accordingly for alleviating the distortion effect. 

\subsection{Direction of CB translation vs. PL deflection}
\label{sec:visualization_sub1}

\begin{figure}[tb]

\begin{minipage}[b]{0.23\linewidth}
  \centering
  \centerline{\includegraphics[width=2.1cm]{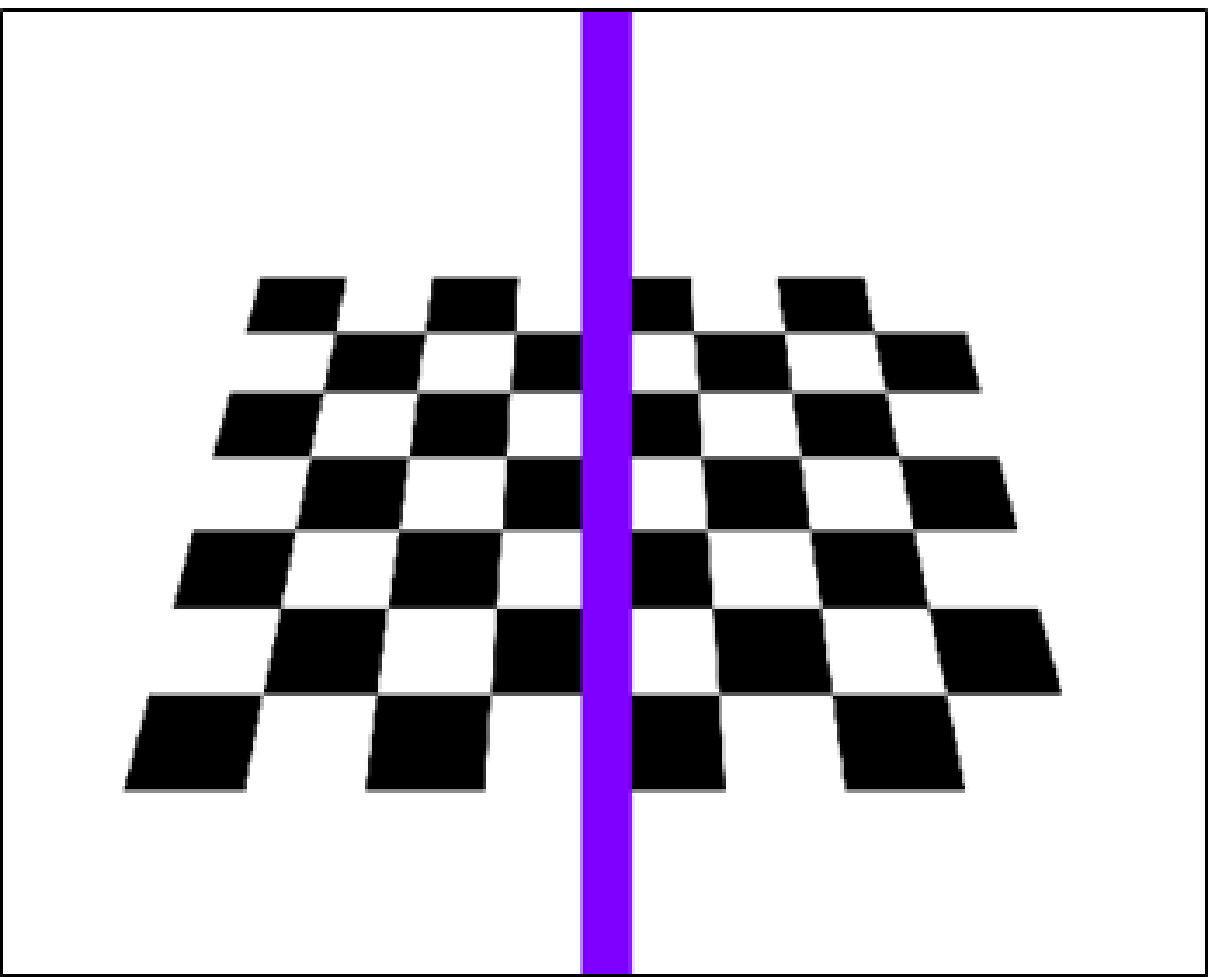}}
  \centerline{(a)}\medskip
\end{minipage}
\hfill
\begin{minipage}[b]{0.23\linewidth}
  \centering
  \centerline{\includegraphics[width=2.1cm]{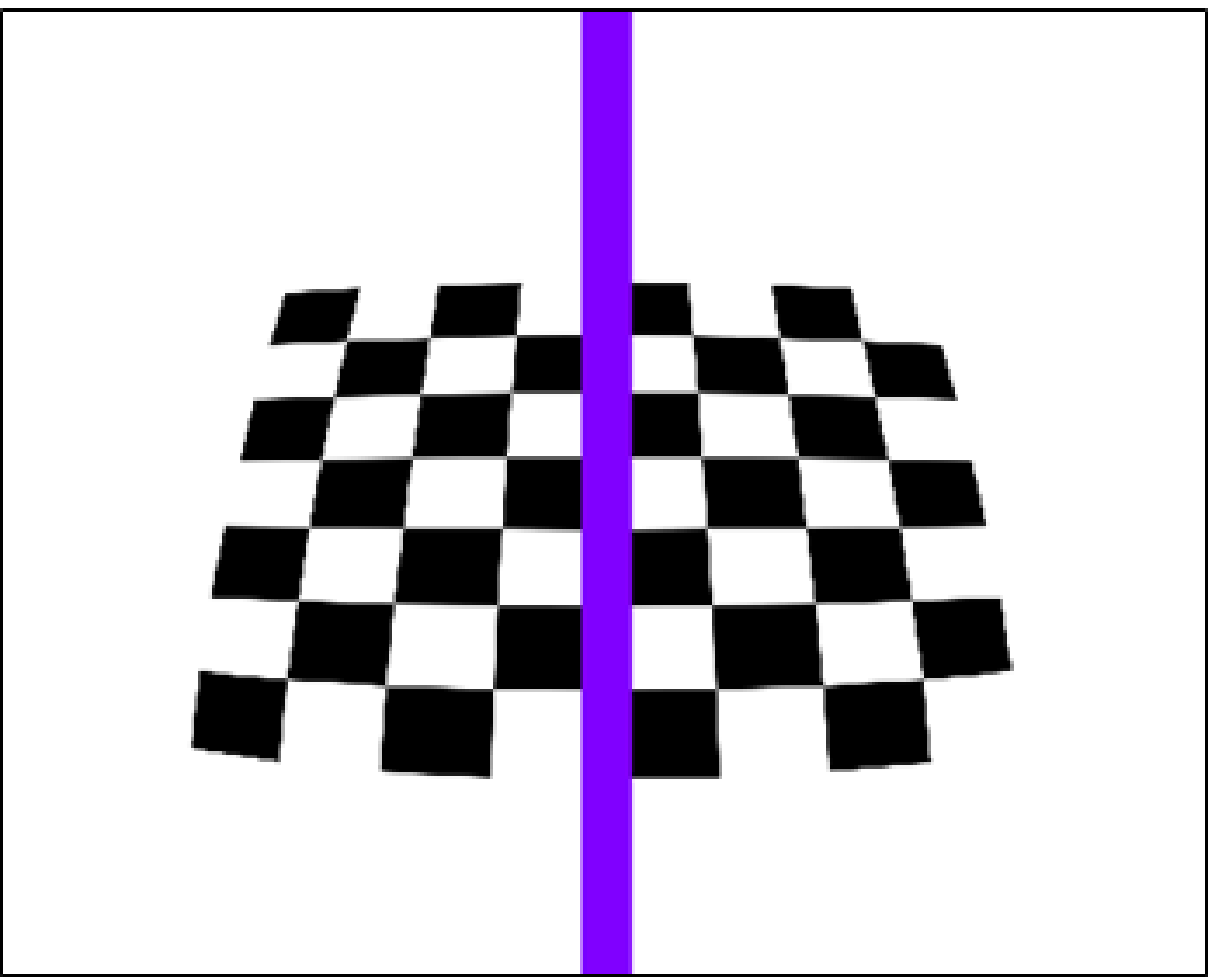}}
  \centerline{(b)}\medskip
\end{minipage}
\hfill
\begin{minipage}[b]{0.23\linewidth}
  \centering
  \centerline{\includegraphics[width=2.1cm]{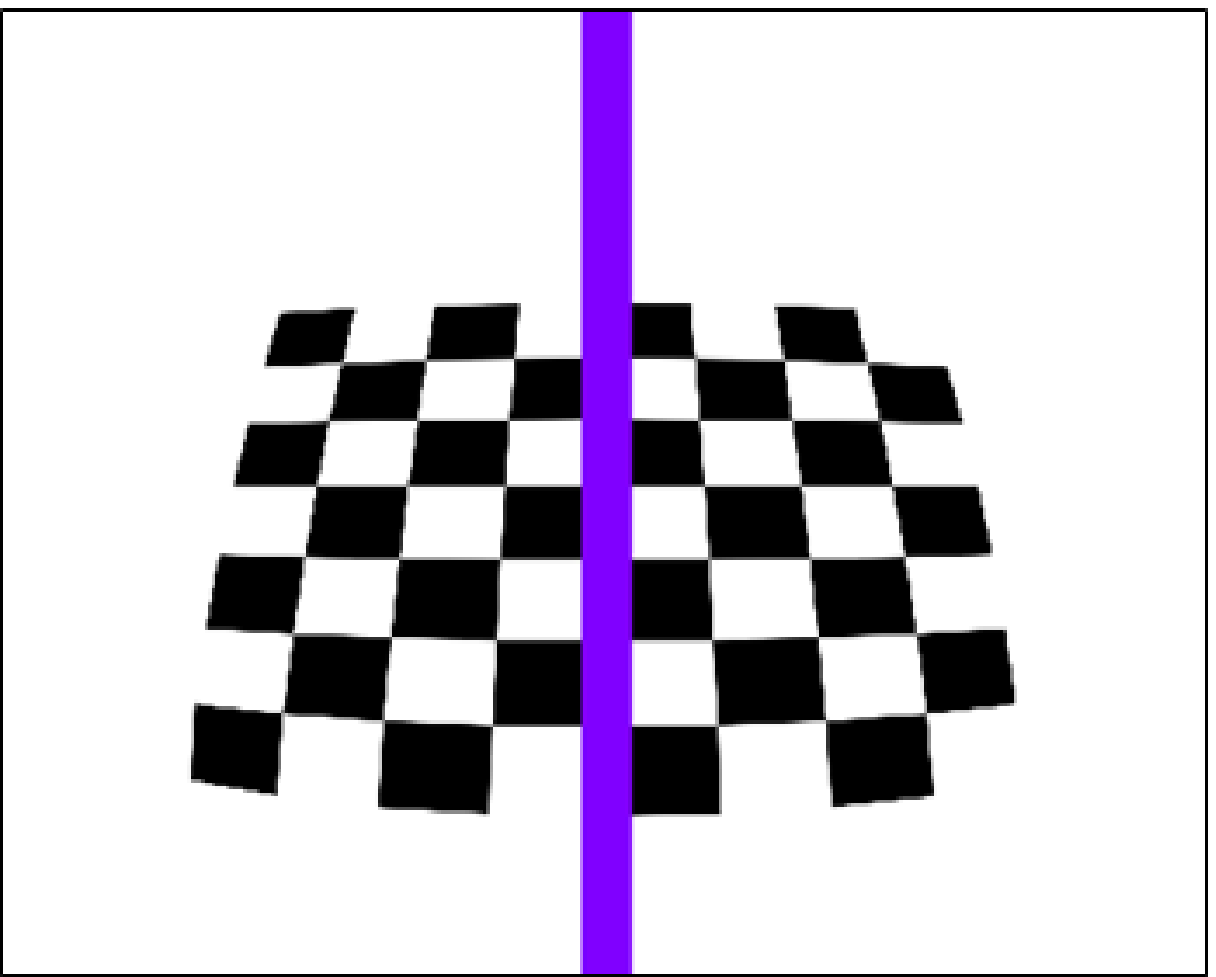}}
  \centerline{(c)}\medskip
\end{minipage}
\hfill
\begin{minipage}[b]{0.23\linewidth}
  \centering
  \centerline{\includegraphics[width=2.1cm]{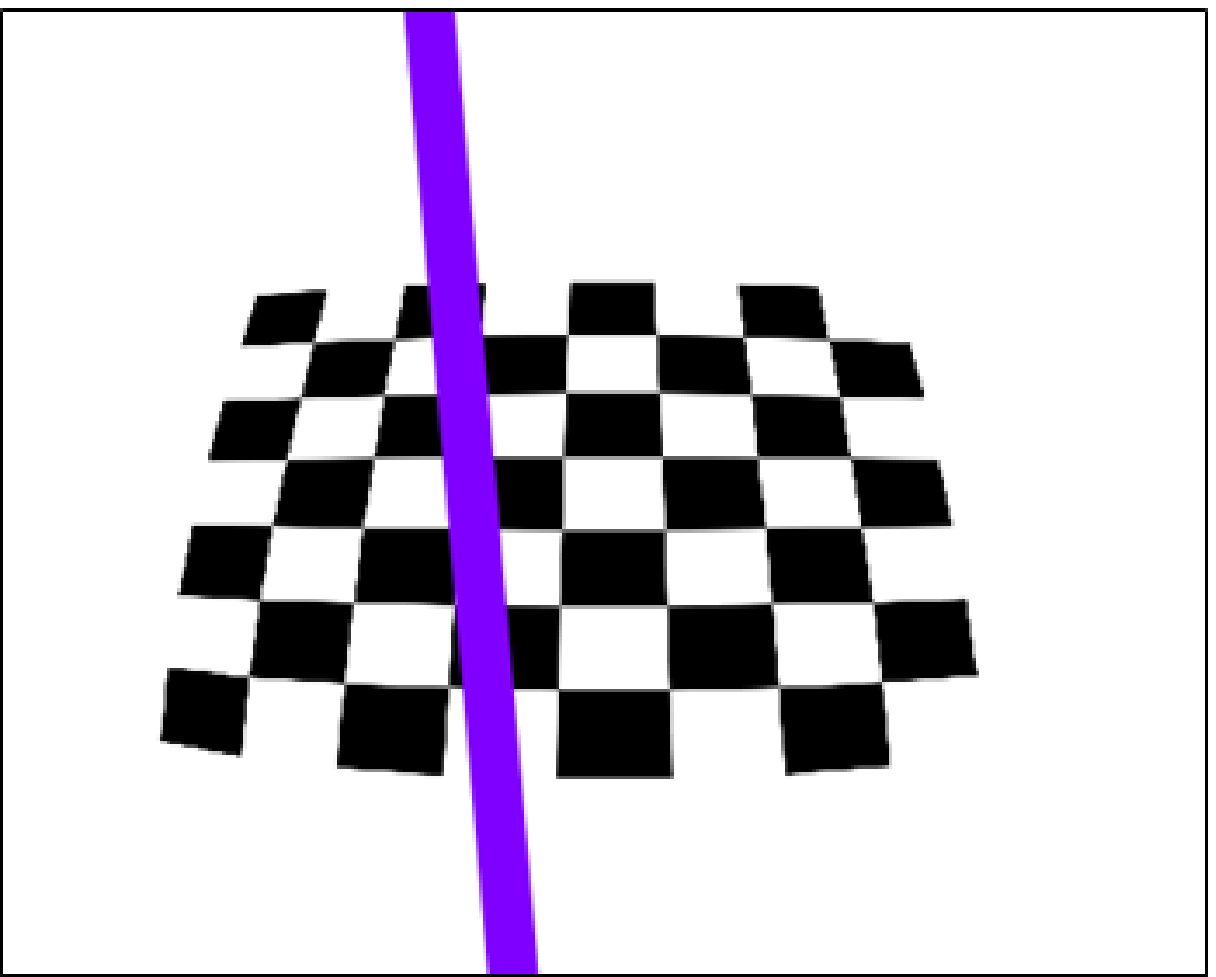}}
  \centerline{(d)}\medskip
\end{minipage}

\caption{(a) PL of the leftmost image shown in Fig. \ref{fig:relationship_of_pimg_and_pchkb} (b), and its distorted versions with CB center translated by (b) (0, 0), (c) (0, 50), and (d) (50, 0) in the \emph{x}-\emph{y} plane.
}
\label{fig:deflected_PL_causedby_distorion}

\end{figure}

Based on the geometry in Fig. \ref{fig:relationship_of_pimg_and_pchkb} (a), the PL (collinear with $y$-axis) is also the axis of symmetry in the camera view for the CB plane, as shown (in purple) in Fig. \ref{fig:deflected_PL_causedby_distorion} (a) wherein a distortion-free and symmetric pattern in the image corresponds to a symmetric pattern on the CB plane. As the deformation of CB due to radial distortion is symmetric w.r.t. PL, the same PL will be estimated, as shown in Fig. \ref{fig:deflected_PL_causedby_distorion} (b). 

Similarly, if the CB in Fig. \ref{fig:deflected_PL_causedby_distorion} (b) is shifted  \emph{radially} along PL, the CB image will still be symmetric w.r.t. the same PL, as shown in Fig. \ref{fig:deflected_PL_causedby_distorion} (c). However, if the same amount of shift is perpendicular to PL, the most distorted CB image will be obtained, resulting in possibly maximal defection of PL, as shown in Fig. \ref{fig:deflected_PL_causedby_distorion} (d).

\subsection{Monotonicity of PL deflection from CB translation}
\label{sec:visualization_sub2}

\begin{figure}[tb]

\begin{minipage}[t]{0.30\linewidth}
  \centering
  \vspace{0pt}
  \centerline{\includegraphics[width=2.1cm]{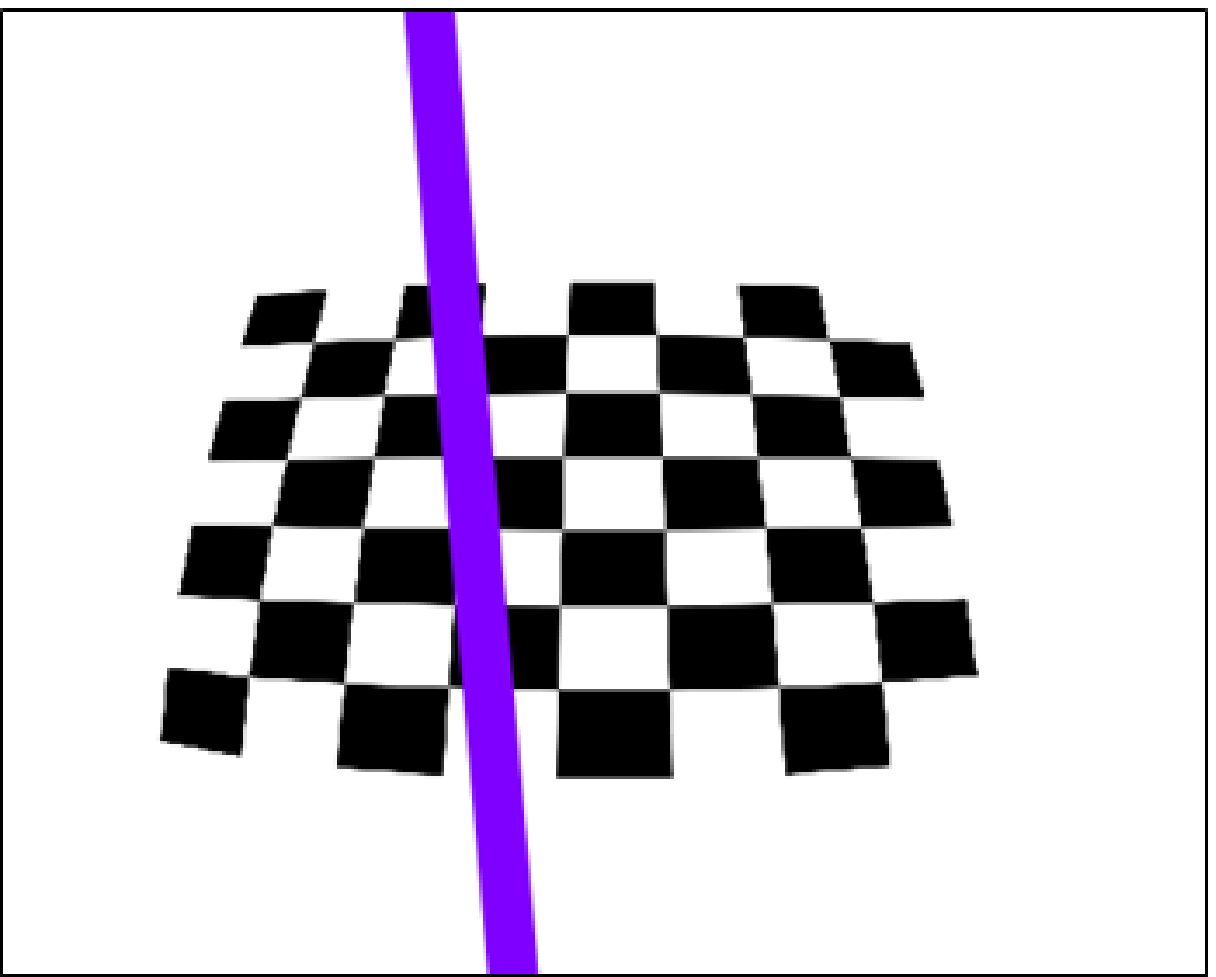}}
  \centerline{(a)}\medskip
\end{minipage}
\hfill
\begin{minipage}[t]{0.30\linewidth}
  \centering
  \vspace{0pt}
  \centerline{\includegraphics[width=2.1cm]{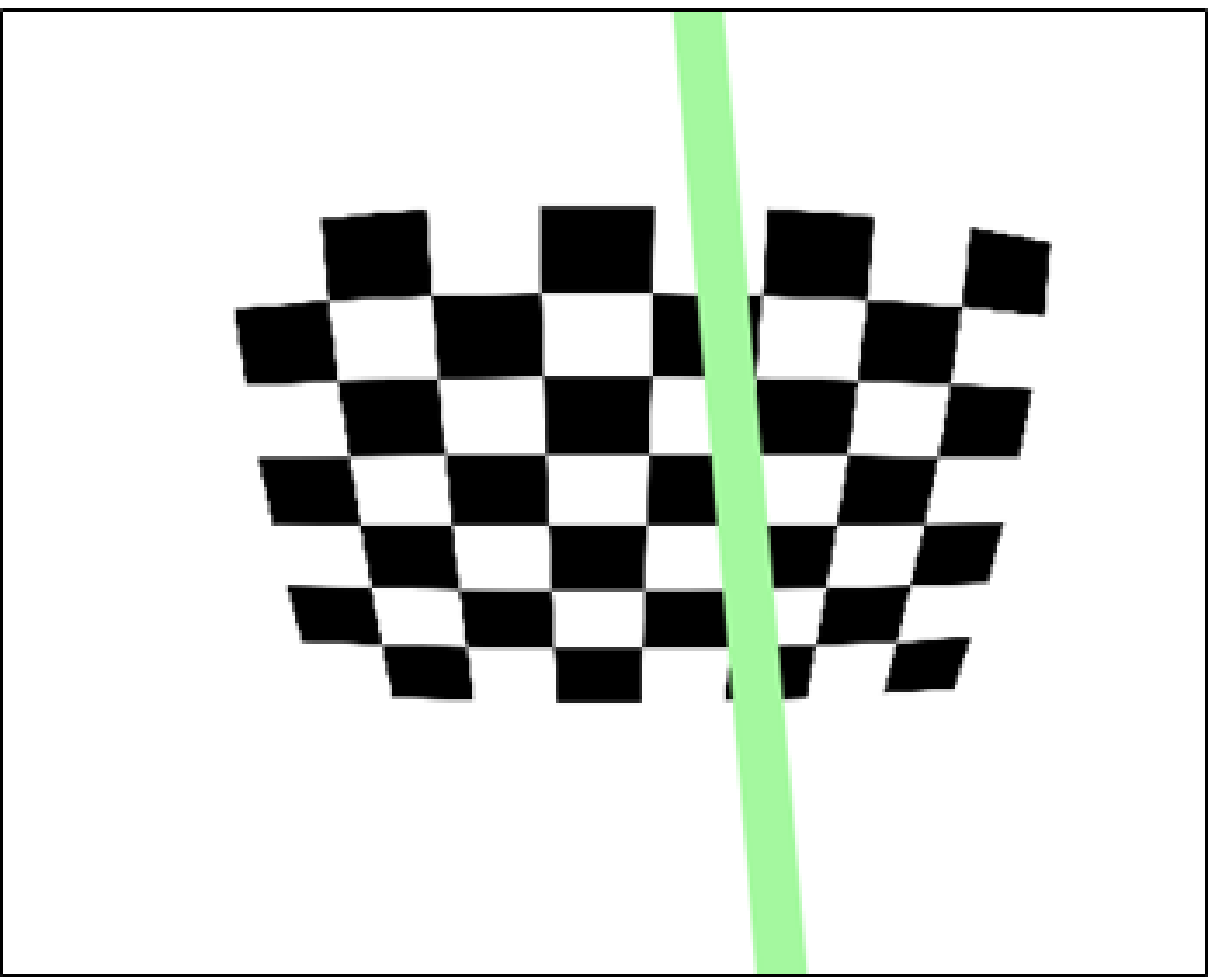}}
  \centerline{(b)}\medskip
\end{minipage}
\hfill
\begin{minipage}[t]{0.30\linewidth}
  \centering
  \vspace{0pt}
  \centerline{\includegraphics[width=2.33cm]{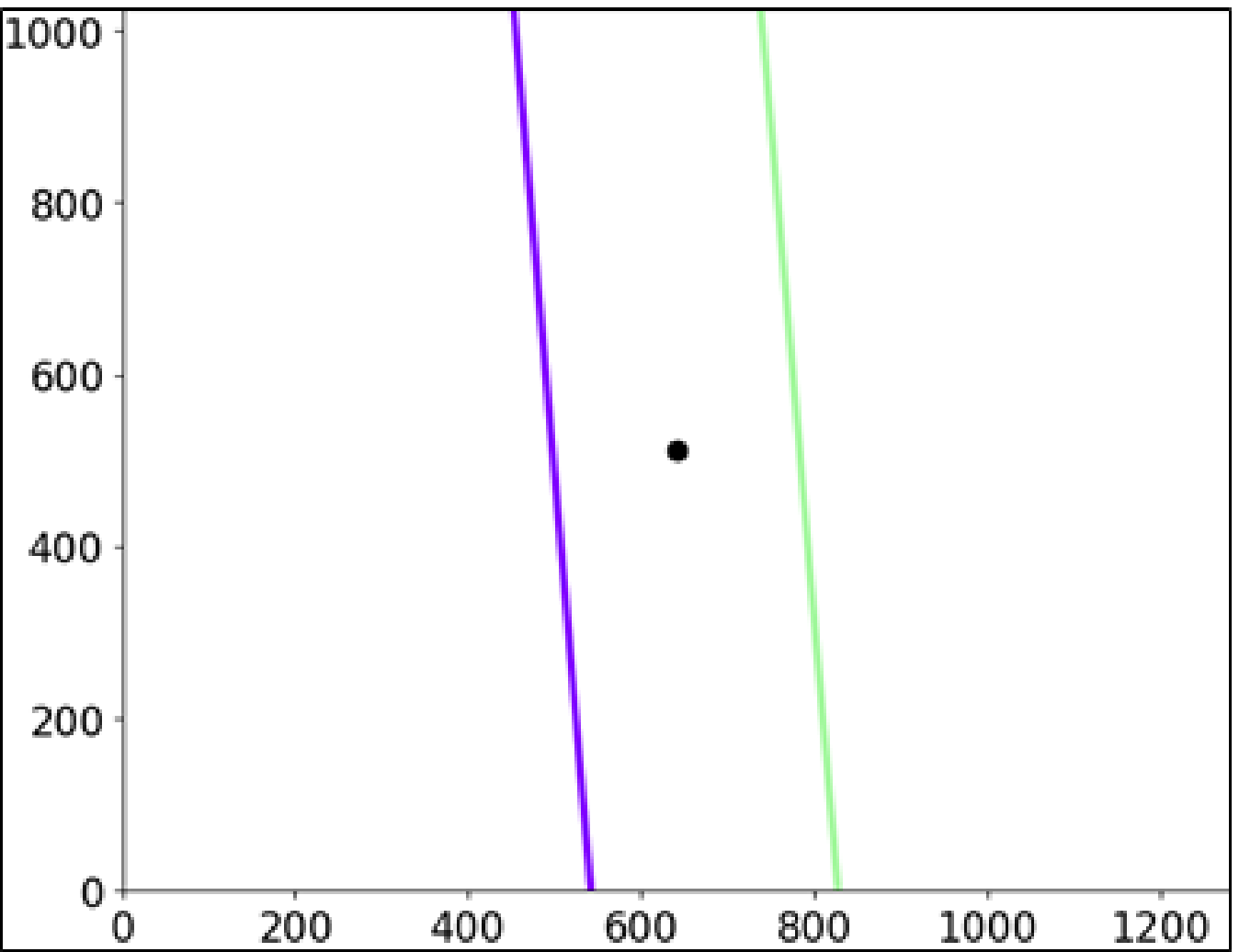}}
  \centerline{(c)}\medskip
\end{minipage}
\begin{minipage}[t]{0.30\linewidth}
  \centering
  \vspace{0pt}
  \centerline{\includegraphics[width=2.1cm]{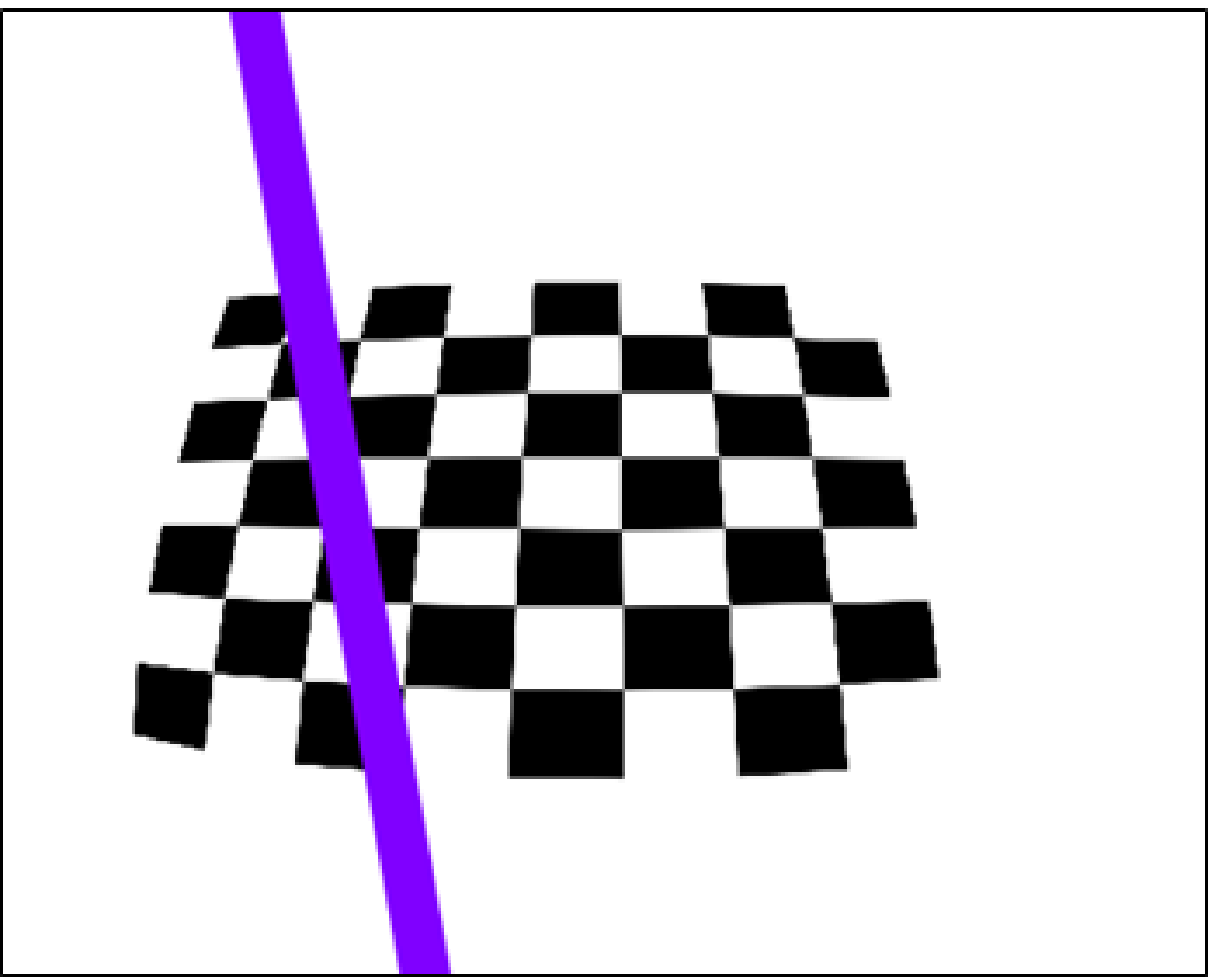}}
  \centerline{(d)}\medskip
\end{minipage}
\hfill
\begin{minipage}[t]{0.30\linewidth}
  \centering
  \vspace{0pt}
  \centerline{\includegraphics[width=2.1cm]{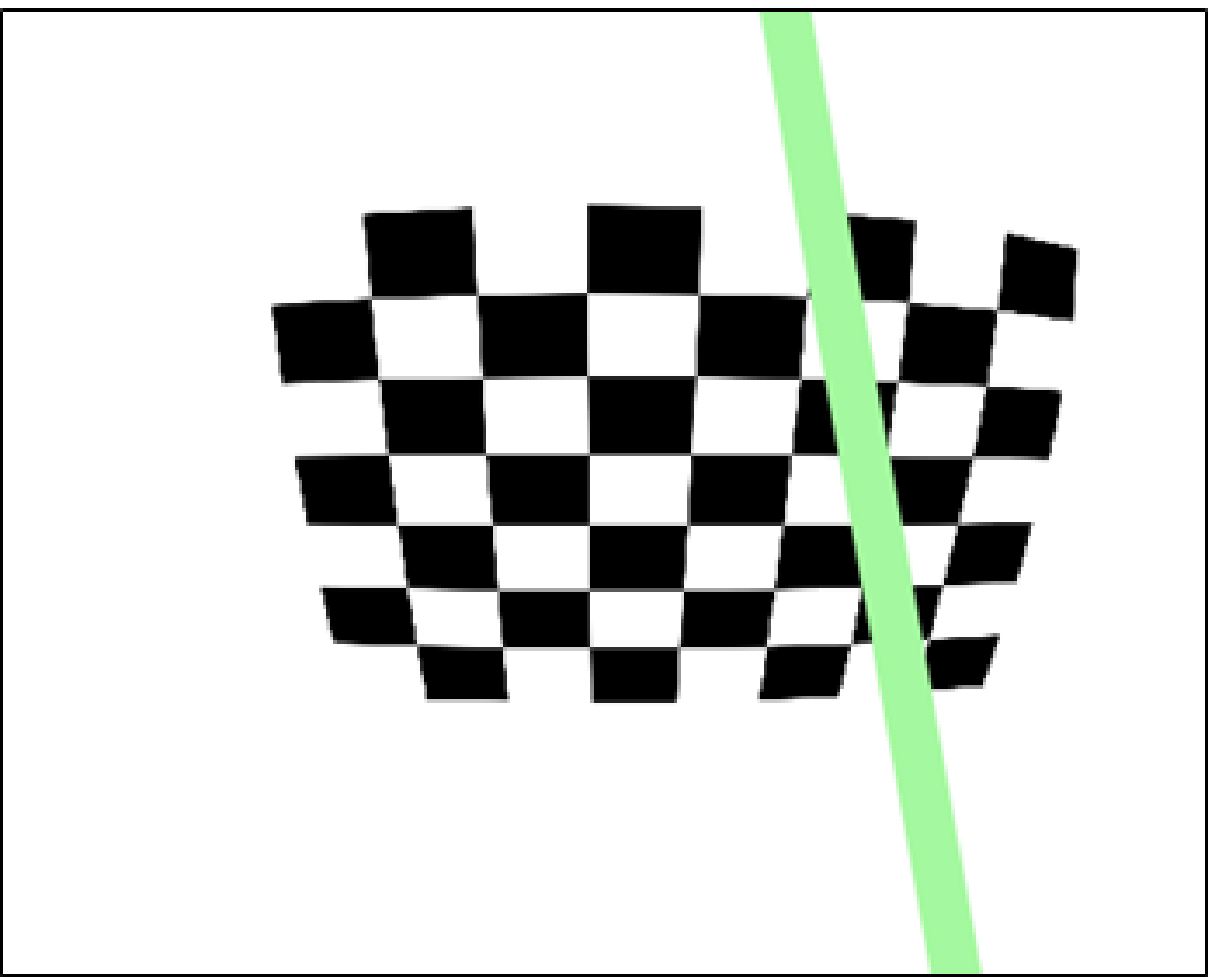}}
  \centerline{(e)}\medskip
\end{minipage}
\hfill
\begin{minipage}[t]{0.30\linewidth}
  \centering
  \vspace{0pt}
  \centerline{\includegraphics[width=2.33cm]{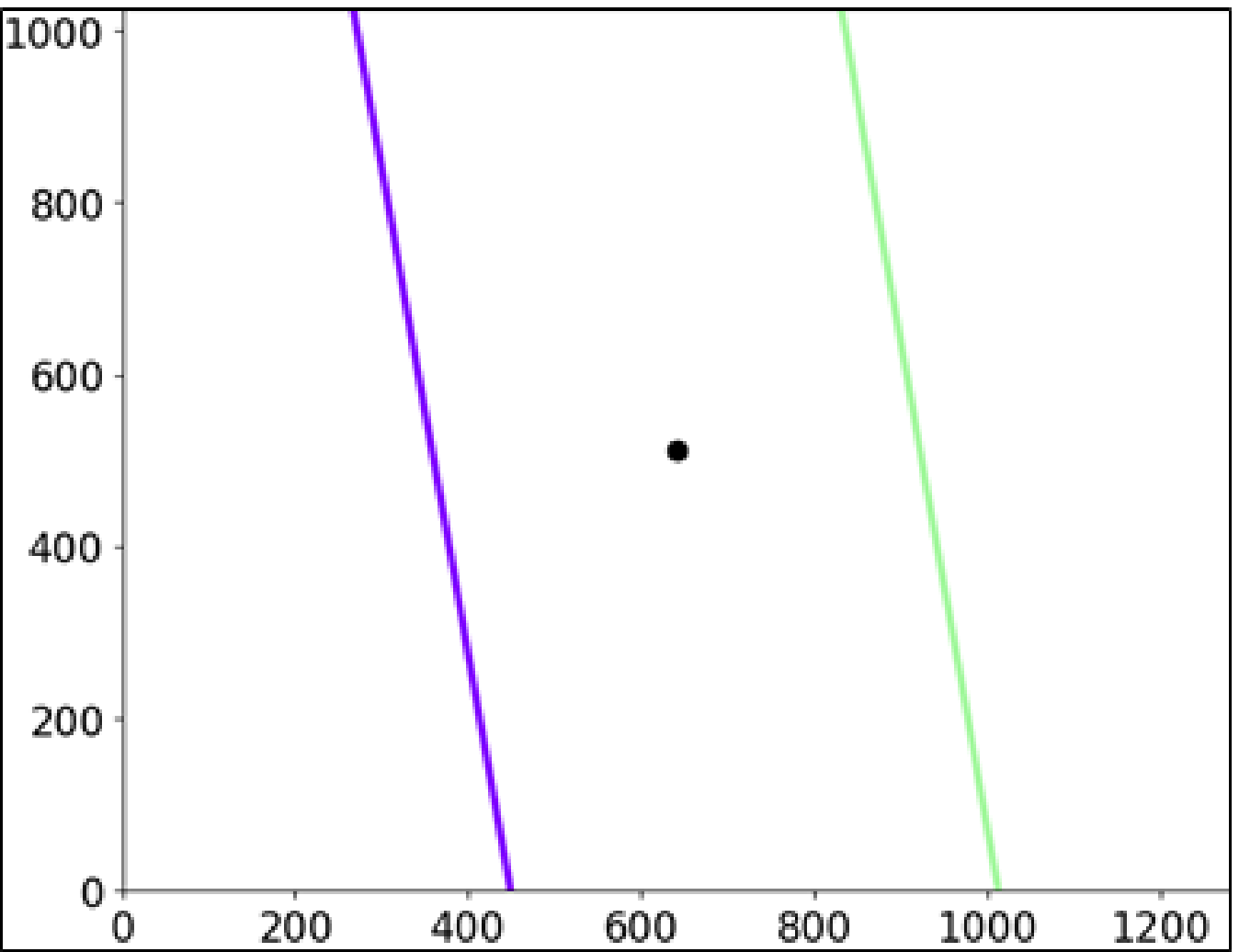}}
  \centerline{(f)}\medskip
\end{minipage}

\caption{(a) PL from Fig. \ref{fig:deflected_PL_causedby_distorion} (d), (b) PL obtained by rotating the CB in (a) by $\alpha=180^\circ$, (c) PLs in (a) and (b). (d)-(f) PLs similar to (a)-(c) but with larger translation. 
}
\label{fig:monotonic_deflected_PL}

\end{figure}

In general, radial distortion bends straight lines more toward the peripheral region; therefore, if the CB image shown in Fig. \ref{fig:deflected_PL_causedby_distorion} (d) (now in Fig. \ref{fig:monotonic_deflected_PL} (a)) is shifted further, more distorted CB image will be obtained, as shown in Fig. \ref{fig:monotonic_deflected_PL} (d). As one may expect, the PL of the latter has more deflection, in both location and orientation, compared with the former. (Nonetheless, a vigorous proof of this is not straightforward as the distortion completely destruct the simple relation of homography between the image plane and the CB plane.)

On the other hand, if we consider the 180° rotated versions of Figs. \ref{fig:monotonic_deflected_PL} (a) and (d), the same amount of PL deflection, but in the opposite direction, will be obtained, as shown Figs. \ref{fig:monotonic_deflected_PL} (b) and (e), respectively. Such results are due to the fact that the deformation is symmetric w.r.t. the rotation. Figs. \ref{fig:monotonic_deflected_PL} (c) and (f) illustrate two pairs of PLs thus obtained, with the two PLs symmetric w.r.t. image center in both figures, showing that effect of distortion may be canceled by using such pairs.

\begin{figure}[b]

\begin{minipage}[b]{0.30\linewidth}
  \centering
  \centerline{\includegraphics[width=2.33cm]{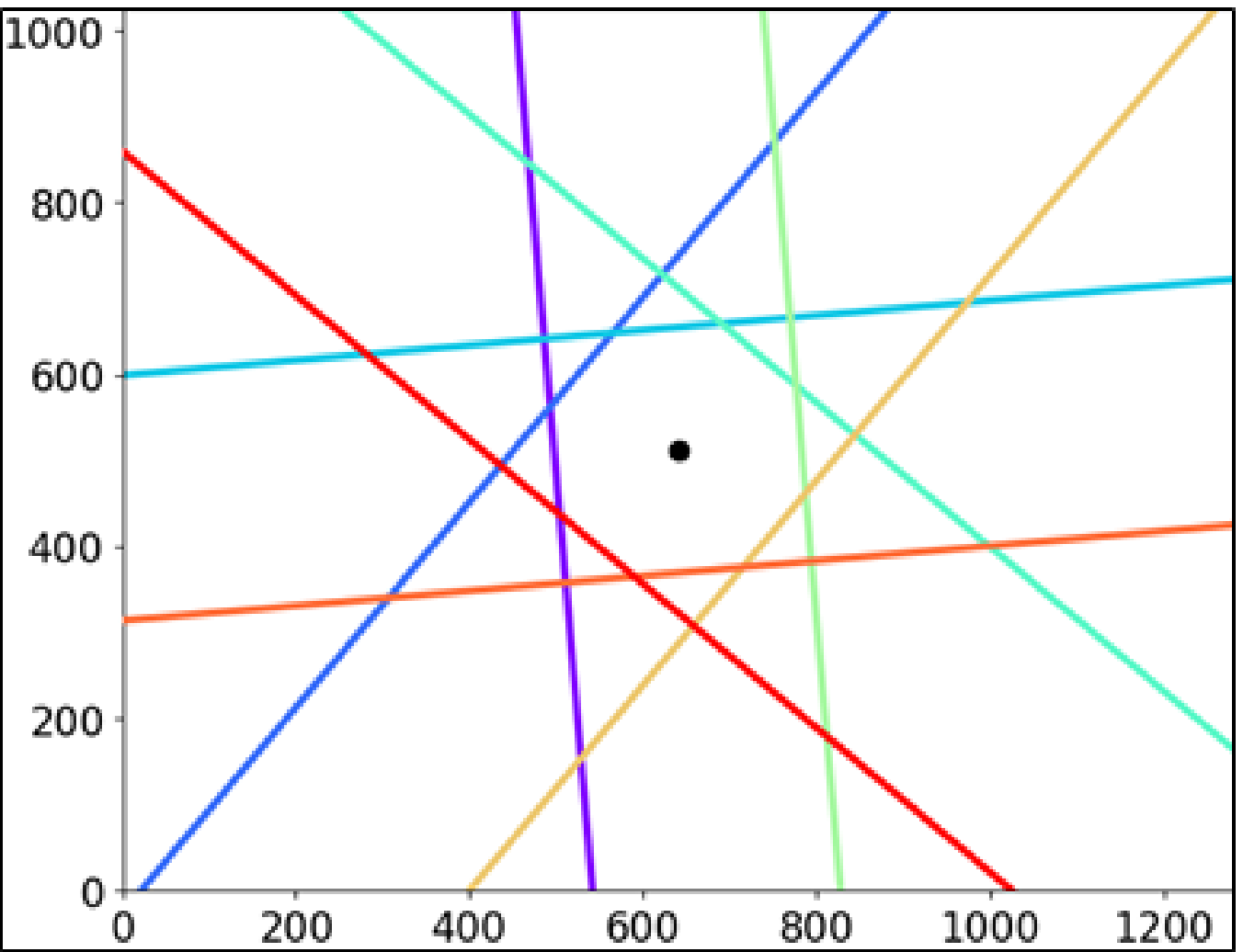}}
  \centerline{(a)}\medskip
\end{minipage}
\hfill
\begin{minipage}[b]{0.30\linewidth}
  \centering
  \centerline{\includegraphics[width=2.33cm]{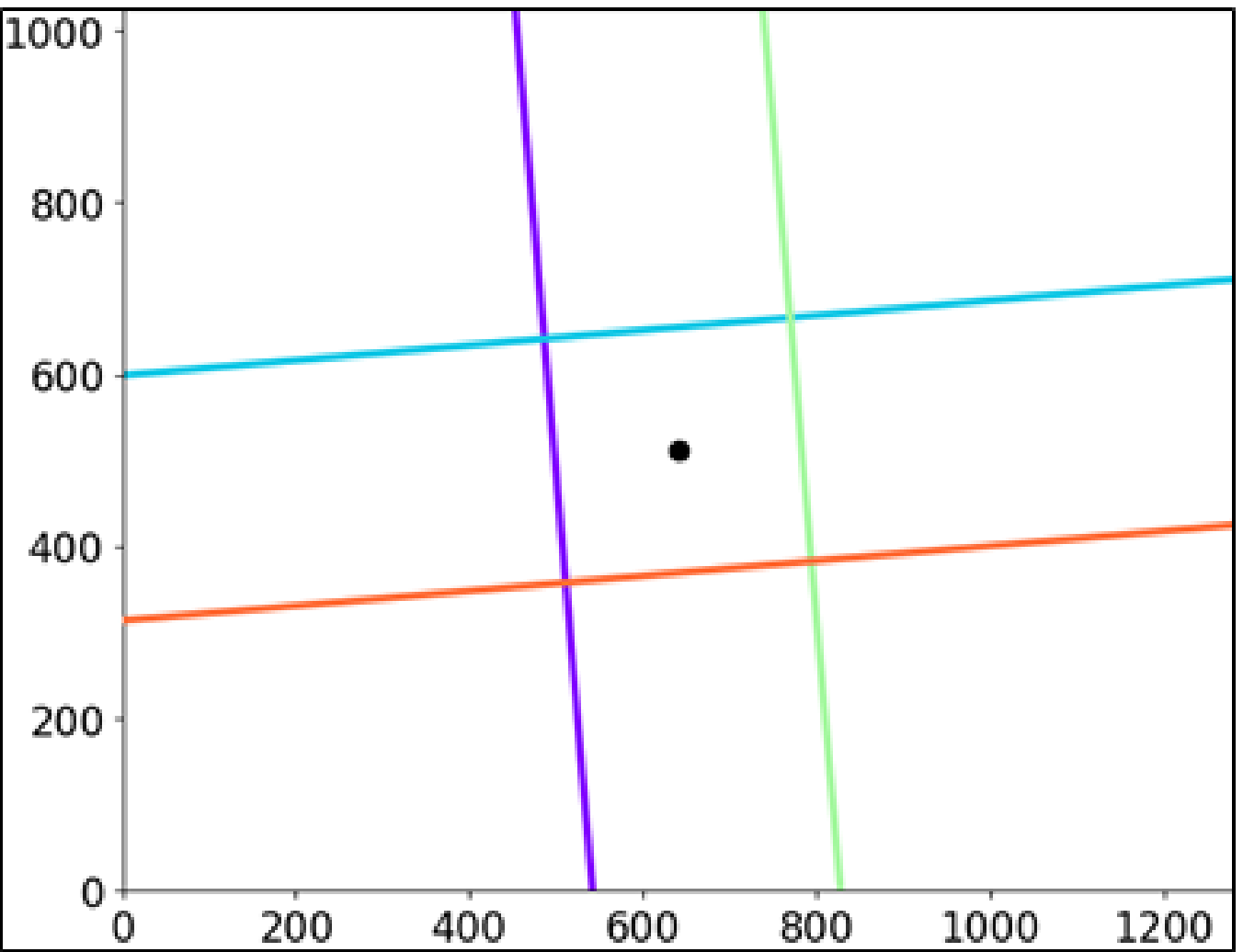}}
  \centerline{(b)}\medskip
\end{minipage}
\hfill
\begin{minipage}[b]{0.30\linewidth}
  \centering
  \centerline{\includegraphics[width=2.33cm]{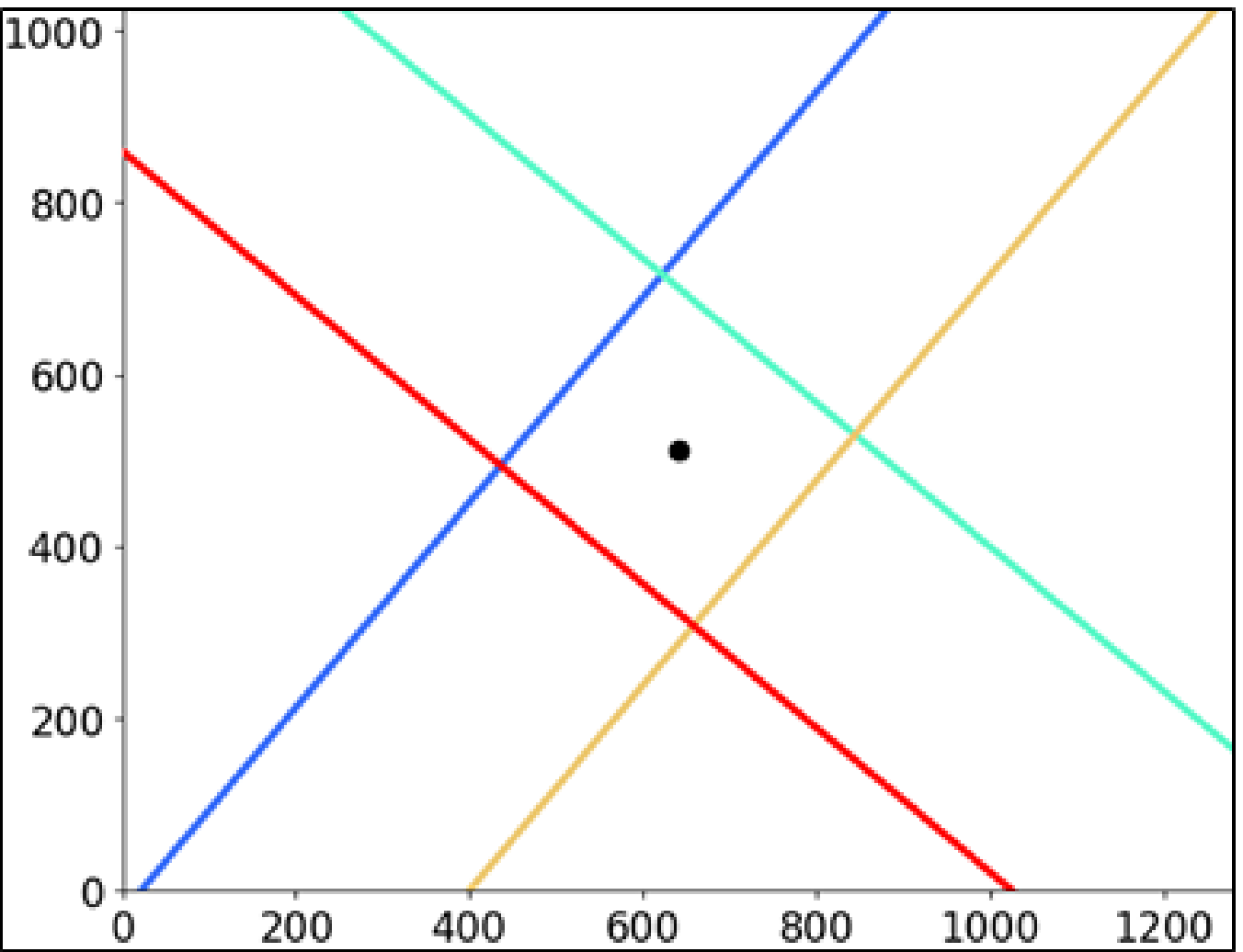}}
  \centerline{(c)}\medskip
\end{minipage}

\caption{(a) Four pairs of PLs (with one PL pair from Fig. \ref{fig:monotonic_deflected_PL} (c)) and the PP estimated according to \cite{chuang2021}; (b) and (c): the same PP obtained with two pairs of PLs (see text). 
}
\label{fig:multiple_deflected_PLs}

\end{figure}

\subsection{Alleviating radial distortion using pairs of PL}
\label{sec:visualization_sub3}

According to the foregoing observations, an effective way of alleviating the effect of radial distortion is to utilize pairs of parallel PLs, from corresponding CBs, like the ones shown in Figs. \ref{fig:monotonic_deflected_PL} (c) and (f). Fig. \ref{fig:multiple_deflected_PLs} (a) shows the eight (four pairs) PLs thus obtained for the geometry shown in Fig. \ref{fig:relationship_of_pimg_and_pchkb}, with the initial translation of (50,  0) for the CB image. Note that for the ideal situation shown in Fig. \ref{fig:relationship_of_pimg_and_pchkb} (a), i.e., the rotation axis and the optical axis ($z$-axis) coincide, the effect of distortion can be eliminated completely, resulting in perfect estimation of image center. 

Ideally, using any two linearly independent pairs of PLs can also achieved excellent accuracy of the estimation, as shown in Figs. \ref{fig:multiple_deflected_PLs} (b) and (c), each using two of the four pairs of PLs shown in Fig. \ref{fig:multiple_deflected_PLs} (a). As for more realistic situations wherein optical axis and image center are unknown during calibration, similar effect can be achieved by using the center of the image sensor, as presented next.

\section{Experimental results}
\label{sec:experiments}

In this section, to explore effective ways of improving camera calibration for distorted CB patterns, experiments are conducted for observations/suggestions presented in Sec. \ref{sec:visualization}. In particular, synthetic data are first employed to verify the directionality/monotonicity mentioned in Sec. \ref{sec:visualization_sub1} and Sec. \ref{sec:visualization_sub2}, while real images are used to demonstrate the improvements in the estimation of PP according to suggestions in Sec. \ref{sec:visualization_sub3}.\footnote{Only PP, not PL deflections, are considered in this section as the latter are not available for comparison for algebra-based methods.}

\subsection{Directionality/monotonicity of effect of distortion}

\begin{figure}[tb]

\begin{minipage}[b]{0.45\linewidth}
  \centering
  \centerline{\includegraphics[width=4cm]{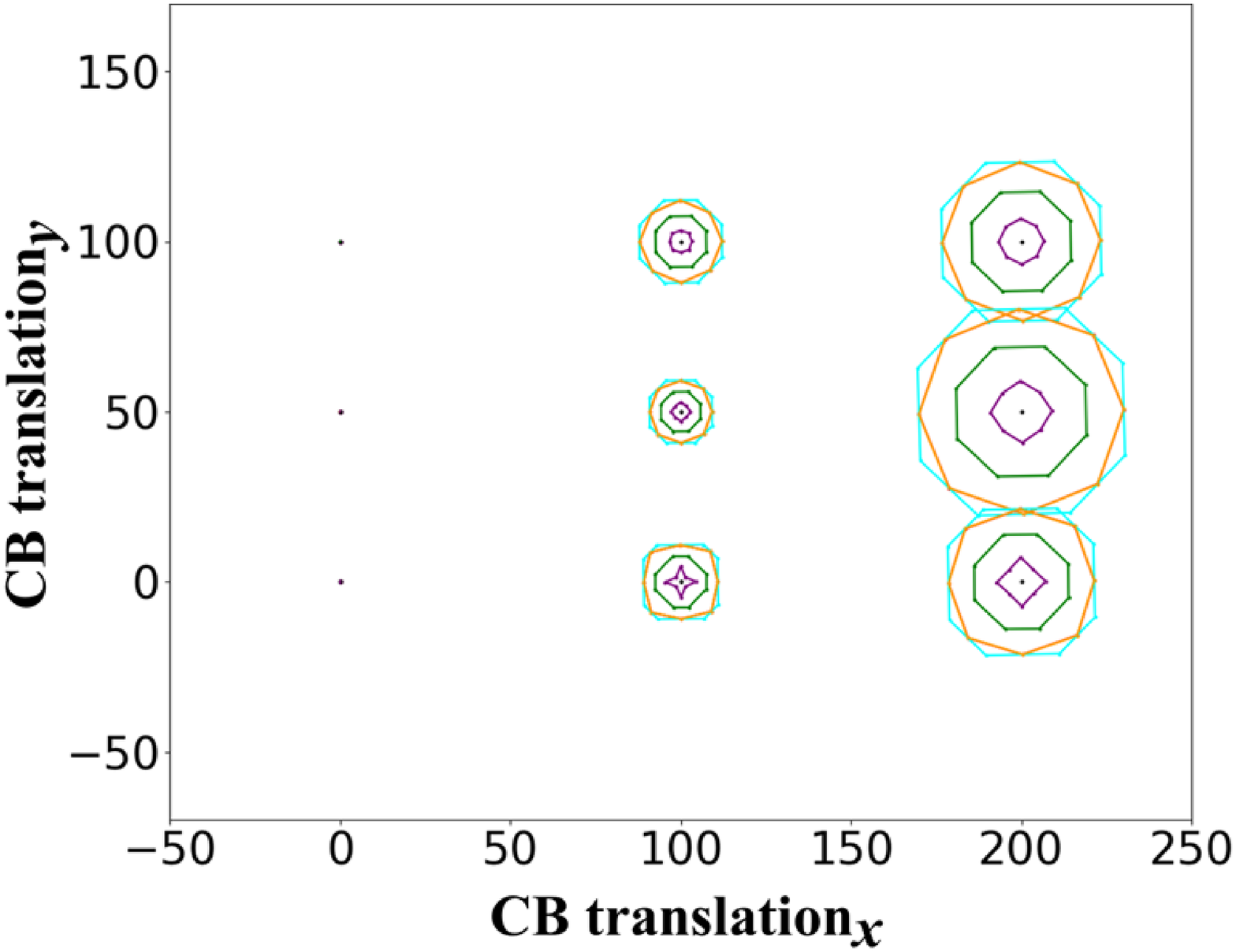}}
  \centerline{(a)}\medskip
\end{minipage}
\hfill
\begin{minipage}[b]{0.45\linewidth}
  \centering
  \centerline{\includegraphics[width=4cm]{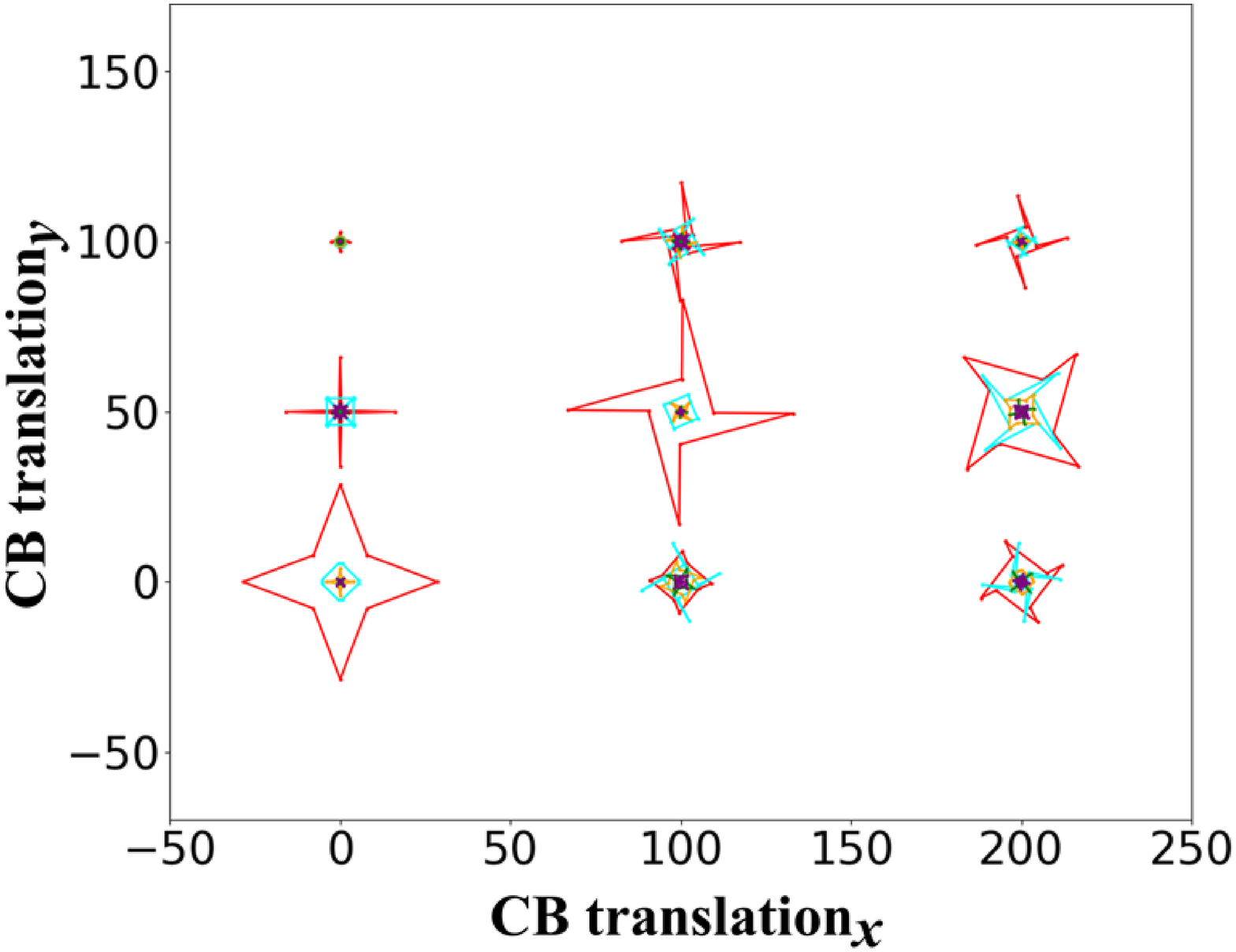}}
  \centerline{(b)}\medskip
\end{minipage}
\vfill
\begin{minipage}[b]{0.45\linewidth}
  \centering
  \centerline{\includegraphics[width=4cm]{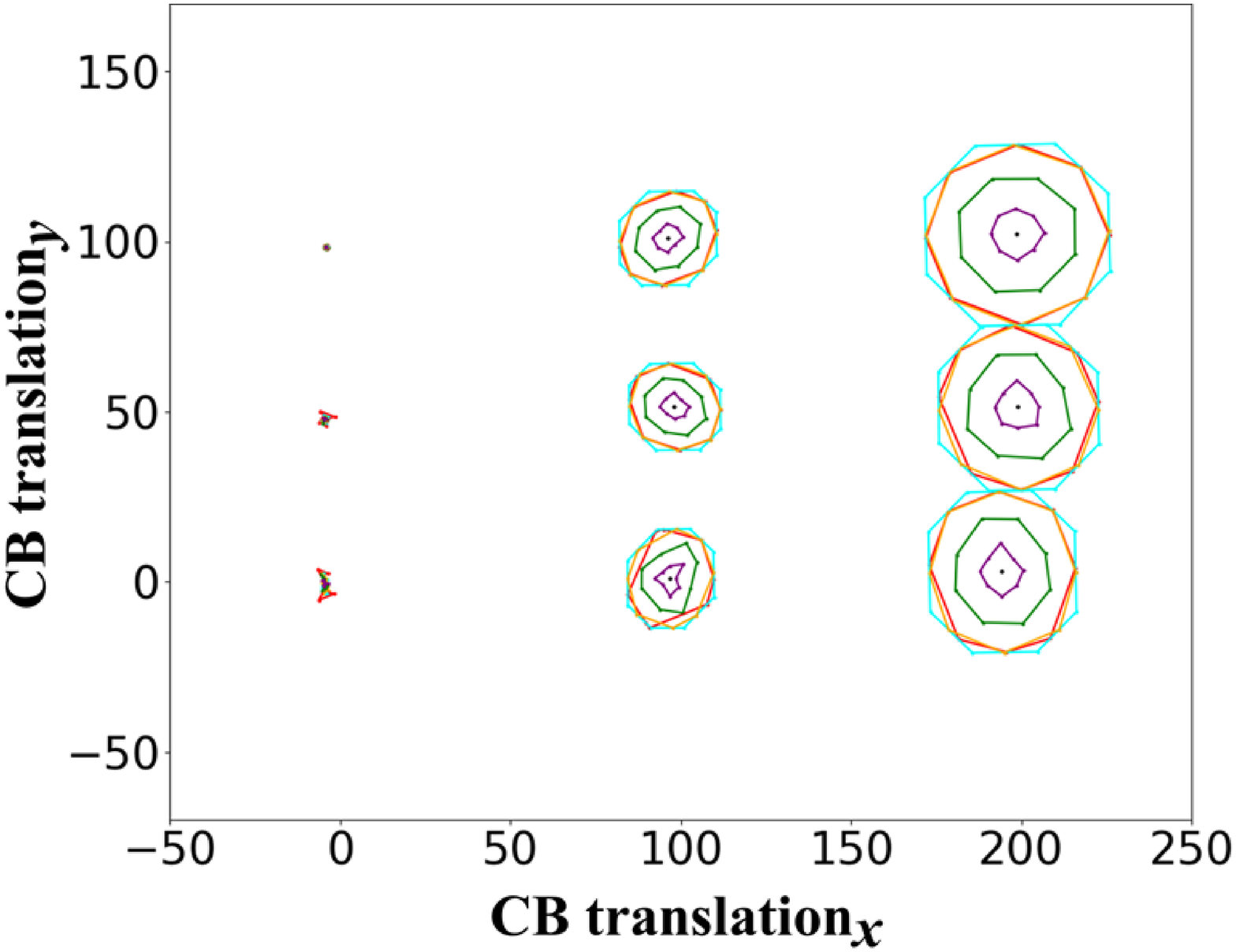}}
  \centerline{(c)}\medskip
\end{minipage}
\hfill
\begin{minipage}[b]{0.45\linewidth}
  \centering
  \centerline{\includegraphics[width=4cm]{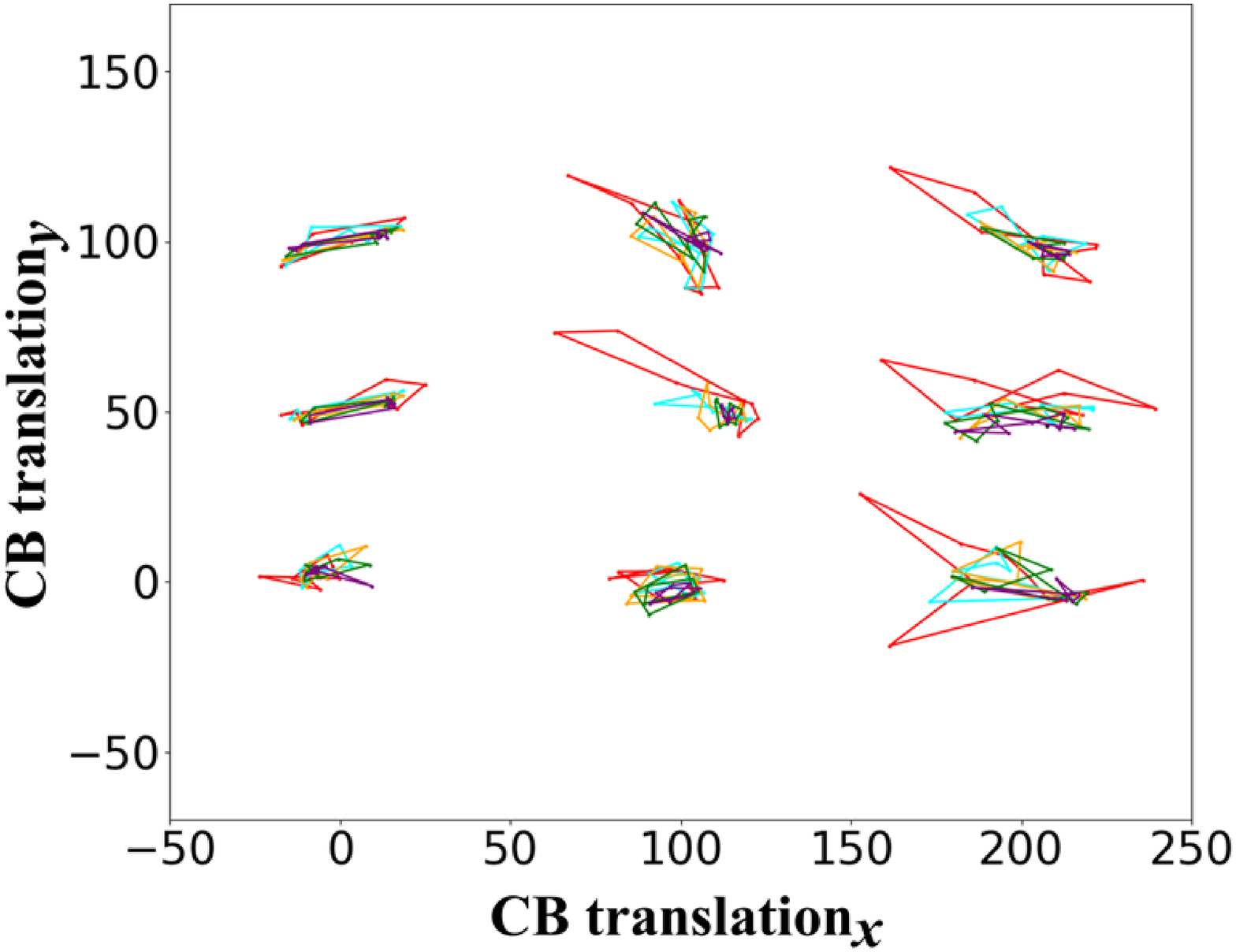}}
  \centerline{(d)}\medskip
\end{minipage}

\caption{Estimations of PP using synthetic data (see text). 
}
\label{fig:_exp_synthetic_distor_effect_by_translation}

\end{figure}

To explore the directionality/monotonicity of the effect of distortion, various CB translations are tested, as shown in Fig. \ref{fig:_exp_synthetic_distor_effect_by_translation}.\footnote{In the experiment settings, virtual camera system having focal length 1600 pixels and distortion parameters $(k_1, k_2)=(-0.1, -0.02)$ is simulated for a CB centered at (0, 0, 2600).} Fig. \ref{fig:_exp_synthetic_distor_effect_by_translation} (a) shows the PP estimations of the proposed approach with different $x$ and $y$ components of CB translation, wherein only the center parts of the image are depicted at the corresponding $x$-$y$ locations. With all eight patterns ($\Delta \alpha = 45^\circ$) used, due to circular symmetry, perfect PPs are obtained for all nine cases. To show the PP drift due to distortion, such symmetry is reduced intentionally by each time skipping one of the eight sets of $n$ consecutive CB patterns (with $n = 1$ to $5$ corresponding to purple, green, orange, cyan, and red, respectively) and estimating a total of eight PPs for each $n$, which are coincide for \emph{CB} $Translations_x = 0$. 

There are several interesting observations for Fig. \ref{fig:_exp_synthetic_distor_effect_by_translation} (a), including: (i) symmetric results are obtained due to image digitization, with odd (and even) $n$'s having identical directions of PP drift, (ii) clear directionality of the amount of drift, with minimum variation in $y$ and more significant variation in $x$ direction, (iii) monotonic trend of the extent of the latter variation in (ii), and (iv) near monotonic trend of the PP drift w.r.t. $n$. As for results from Zhang's method shown in Fig. \ref{fig:_exp_synthetic_distor_effect_by_translation} (b), on the other hand, only (i) and (iv) can be observed. In fact, unlike (iii), small displacement of the CB patterns may actually result in larger PP drifts, which may result from the nonlinear optimization in their PP estimation.

Figs. \ref{fig:_exp_synthetic_distor_effect_by_translation} (c) and (d) shows similar results for both methods but with zero-mean Gaussian noises, with standard deviation 0.5 pixels, added to CB corners. The general trend of PP drift remains the same for these noisy cases except for the symmetry in (i) which seems to be observable from our results but not for Zhang's algebraic approach. Overall, more accurate PP may be obtained with our approach compare to Zhang's method, although the former does not estimate the distortion parameters explicitly like the latter.

\subsection{Robustness of proposed PP estimation}

\begin{figure}[tb]

\begin{minipage}[b]{0.3\linewidth}
  \centering
  \centerline{\includegraphics[width=3.0cm]{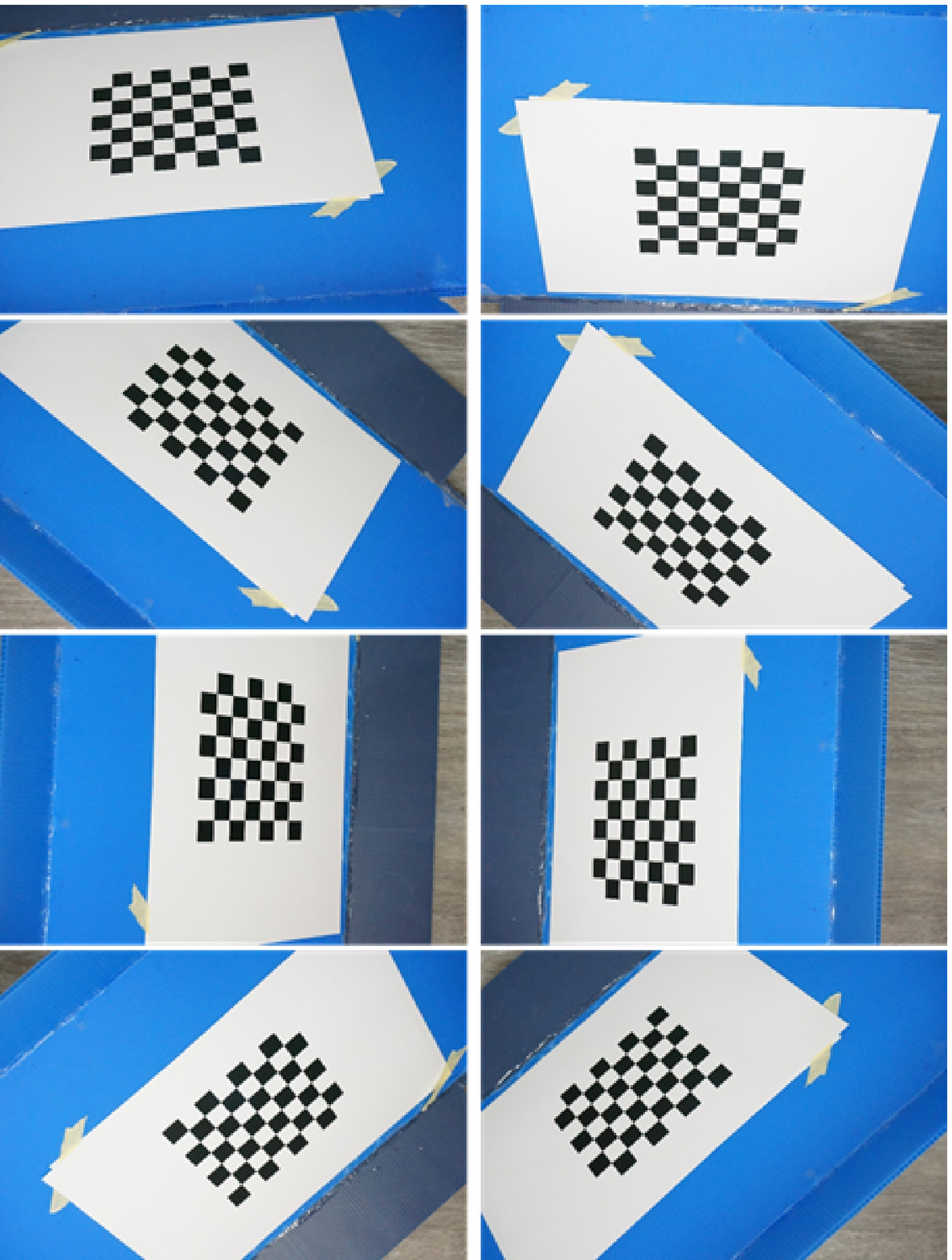}}
  \centerline{(a)}\medskip
\end{minipage}
\hfill
\begin{minipage}[b]{0.65\linewidth}
  \centering
  \centerline{\includegraphics[width=5.0cm]{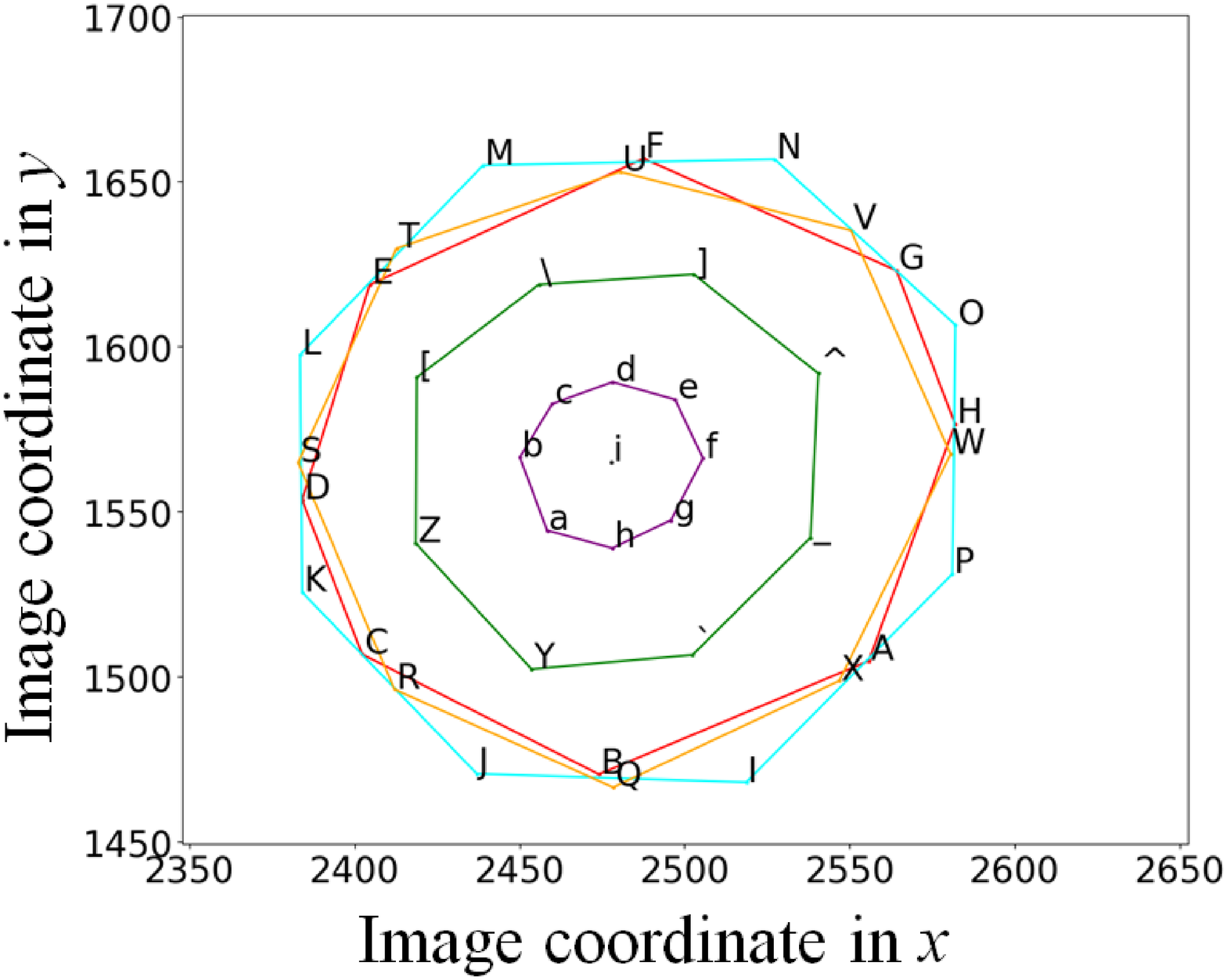}}
  \centerline{(b)}\medskip
\end{minipage}

\caption{(a) Eight CB images captured by a SONY NEX-5R camera. (b) PPs estimated with the proposed approach (see text).
}
\label{fig:_exp_real_fl_40mm_2}

\end{figure}

To show the robustness of proposed PP estimation, eight 4912 x 3264 CB images, as shown in Fig \ref{fig:_exp_real_fl_40mm_2} (a), are captured by  a SONY NEX-5R camera with manual and approximate alignment according to the geometry shown in Fig. \ref{fig:relationship_of_pimg_and_pchkb} (a) without using any special equipment. Specifically, by roughly maintaining the dihedral angle in Fig. \ref{fig:relationship_of_pimg_and_pchkb} (a) to $45^\circ$, rotations for obtaining these images are performed with an off-center CB corner fixed at (2956, 1632). Fig. \ref{fig:_exp_real_fl_40mm_2} (b) shows various PP estimations obtained with the proposed approach, which also have near circularly symmetric distributions (color groups) similar to those depicted in Fig. \ref{fig:_exp_synthetic_distor_effect_by_translation} (a).\footnote{Distributions of more irregular shapes can also be obtained for Zhang's method, as one may expect, which are not displayed here for brevity.} Such distributions are expected to alleviate the effect of radial distortion effectively if centroids of all groups are close to one another, as we may see next with comparisons to Zhang's method. 

\begin{table}[tb]
\small \addtolength{\tabcolsep}{-2.2pt}
\centering
\caption{
Performance evaluation for distortion alleviation. \\
}
\label{tab:_exp_real_fl_40mm_2}
\begin{threeparttable}
\resizebox{1.0\hsize}{!}{
\begin{tabular}{|c|c|c||c|c|}
\hline
\multirow{2}{3cm}{(8-$n$)-image sets for Fig. 6's PP estimates} & \multicolumn{2}{c|}{Centroid of PPs (Ours)} & \multicolumn{2}{c|}{Centroid of PPs (Zhang's)} \\ \cline{2-5}
 & $x$ & $y$ & $x$ & $y$ \\ \hline
\makebox[3cm][l]{$n=5$ (group of red)} &	2481.85 & 1563.81 & 2440.70 & 1550.50 \\ \hline 
\makebox[3cm][l]{$n=4$ (group of cyan)} &	2481.62 & 1563.85 &	2408.00 & 1639.34 \\ \hline 
\makebox[3cm][l]{$n=3$ (group of orange)} &	2480.57 & 1563.90 &	2368.99 & 1587.62 \\ \hline 
\makebox[3cm][l]{$n=2$ (group of green)} &	2478.86 & 1564.29 &	2358.22 & 1592.15 \\ \hline 
\makebox[3cm][l]{$n=1$ (group of purple)} &	2477.93 & 1564.85 &	2354.64 & 1591.03 \\ \hline 
\makebox[3cm][l]{$n=0$ (one black PP)} &	2477.76 & 1564.99 &	2356.40 & 1592.08 \\ \hline 
\makecell{Mean / Std (all $n$'s)} & \makecell{2479.76 / \textbf{1.82}} & \makecell{1564.28 / \textbf{0.52}} & \makecell{2381.16 / 35.37} & \makecell{1592.12 / 28.22} \\ \hline 
\makecell{Mean / Std (9 test sets)} & \makecell{2481.83 / \textbf{2.01}} & \makecell{1568.07 / \textbf{1.37}} & \makecell{2380.21 / 36.73} & \makecell{1579.27 / 25.66} \\ \hline 

\end{tabular}
}
\end{threeparttable}
\end{table}

Table \ref{tab:_exp_real_fl_40mm_2} shows centroid of eight PPs estimated for each color group shown in Fig. \ref{fig:_exp_real_fl_40mm_2} (b), with the last group ($n=0$) having just one PP, followed by the mean and standard deviation of all centroids. Important observations of these results include: (a) the two means of the six centroids are very similar for both methods (less than 30 pixels apart) and (b) the two standard deviations for (a) is quite different (having a 20+ ratio). While the validity of the experiment can be clearly confirmed with (a), our approach seems to be more robust in alleviating the effect of distortion with (b). In addition, the two means are about 24 and 2 pixels away from the ordinary PP estimates ($n=0$) for Zhang's and our methods, respectively, showing the consistency of our estimation of PP. More test sets are also included in the experiment\footnote{The image data were previously used in \cite{chloe_thesis} where intriguing patterns similar to Fig. \ref{fig:_exp_real_fl_40mm_2} (b) are first observed.}, as shown in the last row of Table \ref{tab:_exp_real_fl_40mm_2}, to reconfirm the foregoing observations, and demonstrate the superior ability of the proposed approach in coping with image distortion.

\section{Conclusion}
\label{sec:conclusion}

In this paper, we demonstrate both directionality and monotonicity of the effect of radial distortion, with respect to CB translations in the image plane, by using deflection of PL as a tool for better visualization. Accordingly, suggestions of using linearly independent pairs of parallel PLs, i.e., using a CB image and its $180^\circ$ rotated version in each pair, are proposed to alleviate such effect. Experimental results show that more robust and consistent calibration results for image center can actually be obtained with the proposed approach, compared with the renowned algebraic methods which estimate distortion parameters explicitly. Other ways of improving camera
calibration, based on analysis of PL variations, are currently under investigation.

\bibliographystyle{IEEEbib}
\bibliography{strings,refs}

\end{document}